\documentclass[twoside]{article}
\usepackage{ukai}

\title{\customtitle{\shortstack{Enhancing Transformer Representations of Symbolic\\
ODE Expressions}}}

\setauthorsshort{Fan et al.}

\author{
    Xiyue Fan$^1$, Adam Prugel-Bennett $^1$, Stuart E. Middleton$^1$ \\
    {$^1$University of Southampton} \\
    \texttt{X.Fan@soton.ac.uk, apb@ecs.soton.ac.uk, sem03@soton.ac.uk}
}

\date{}

\begin{document}
\maketitle

\begin{abstract}
Existing approaches to solving differential equations, such as symbolic regression, physics-informed neural networks, and neural operators, typically focus on numerical approximations or blind symbolic search via fitting to numerical data. Less attention has been paid to learning structured representations of mathematical expressions that preserve commutative properties and could support mathematical reasoning in symbolic forms. Transformer models have shown strong capabilities in solving symbolic differential equations. However, standard positional embeddings in transformers are designed for sequence data. Symbolic differential equations are naturally represented by expression trees, so these positional embeddings may not efficiently capture their hierarchical structures. We investigate existing tree positional embeddings in symbolic ordinary differential equation (ODE) tasks. We systematically study their effectiveness under different settings. Our results show that tree positional embeddings aid learning in early epochs and continue to improve performance throughout, ultimately yielding consistent advantages across various data sizes and tasks. Based on learned structural representations, we apply contrastive learning to support the commutative property in mathematics. Ablation studies provide insight into how these methods interact in modelling symbolic mathematical structures. 
\end{abstract}


\section{Introduction}
Transformer-based models, especially Large Language Models (LLMs), have shown remarkable progress in performing mathematical tasks. They are the dominant models in symbolic regression \cite{LLMSR, 65, SR12}, physics-informed neural networks \cite{76, PLLM, PhysicsSolver}, and neural operators \cite{5, Transolver}. Even though they have achieved substantial progress in different areas in terms of generalisation, accuracy and computation efficiency \cite{5, Transolver, GAOT2026, 76, PLLM, PhysicsSolver}, physics-informed neural networks and neural operators lack interpretability, while symbolic regression suffers from low search efficiency in the symbolic candidate space \cite{5, SR8, Li2023}. 
In contrast to these numerical approaches, evidence shows that training a neural network on symbolic data can enhance interpretability and increase model scalability and generalisation \cite{X9}. Lample and Charton \cite{ref} used an encoder-decoder transformer to process symbolic data and demonstrated its effectiveness in solving differential equations (DEs), finding solutions to many problems that Mathematica was unable to solve under a fixed time budget. However, they treated these symbolic DE expressions as sequential data. The dominant sequential-based positional embeddings in transformers do not match the structures of symbolic differential equations, and thus may not fully encode their rich positional information. These mathematical expressions are naturally represented as tree structures. We hypothesise that injecting this tree-structured hierarchy into a transformer would result in better representations. These structures encode parent-child and sibling relationships among nodes, delineate the functional scope of subtrees, and may facilitate identification of similar substructures. 

Existing studies have demonstrated that transformer models benefit from incorporating tree positional information through embeddings comparing sequence positional embedding in aiding Computer Algebra Systems \cite{TC} or code generation task \cite{Barket}. Even though there is no standard technique to guide this implementation. Tree positional embeddings have shown promise in code translation, mathematical retrieval, and symbolic DE tasks \cite{X31, X28, Scar6Tree, SymPlex}. Recent work has incorporated structural information into symbolic DE solvers through tree-relative self-attention \cite{SymPlex}. While their approach modifies attention using structural information, the effectiveness of tree positional embeddings at the input-representation level remains largely unexplored in the context of symbolic differential equations. 

One issue with either absolute positional embeddings or tree positional embeddings is that they treat formulae that differ by commuting the arguments of commutative operators as different expressions.  For example, the equations
\begin{align}
    \frac{\mathrm{d} y}{\mathrm{d} x} &= \frac{1}{1 + 2\,x},
    &
    \frac{\mathrm{d} y}{\mathrm{d} x} &= \frac{1}{2\,x + 1},
    &
    \frac{\mathrm{d} y}{\mathrm{d} x} &= \frac{1}{x\,2 + 1}
\end{align}
would each have different positional embeddings, despite being mathematically equivalent and having the same solution.  To encourage the model to learn this equivalence, we have used contrastive learning to decrease the distance in latent space between formulae that are mathematical equivalent.  In particular, we consider all nodes of the expression tree that correspond to the addition or multiplication operators and exchange the left and right subtrees.  In contrastive learning we used a loss function that minimised the distance between these pairs of formulae, while maximising the distance between expressions that represent different equations.  The hope was that if the model could identify mathematically identical expressions, this would reduce the vast combinatorial growth of valid mathematical expressions that is common recognised as an obstacle to complex mathematical reasoning tasks \cite{SR8,21}.  Previous work has integrated contrastive learning into LLMs to enhance their mathematical reasoning abilities, but none has investigated its utility for symbolic ODEs. Some works have attempted to learn algebraic properties (i.e., the commutative and identity properties) via supervised training on various data \cite{X30, Chang2024}.  As we will see, the improvements of using contrastive learning for our task are minimal (and in some cases counterproductive). Instead, we get better performance by augmenting the training data by randomly permuting arguments of the addition and multiplication operator.

This paper expands on the work of Lample and Charlton~\cite{ref} in two directions:
\begin{itemize}
    \item We investigated the utility of a transformer model with tree positional encoding for solving symbolic ODE problems, demonstrating that the structural representation improves performance in modelling symbolic ODE expressions.
    \item We examined how well contrastive learning captures the commutative property. Our findings indicate that prior structural representations are essential for learning the commutative property. However, the representation invariance acquired through this process does not provide additional benefits over direct supervised augmentation training.
\end{itemize}

\section{Approach}

As a baseline we have used the model of \cite{ref}.  This uses a vanilla encoder-decoder transformer model \cite{X25}, but trained it on differential equations and their solutions. The author achieved very strong results, finding solutions to problems that Mathematica were unable to solve under a fixed time budget. In their work both the differential equations and solutions were written in Polish notation and treated as a string, using standard absolute positional embeddings. In this paper we explore two extensions: the use of tree embeddings, and the use of contrastive learning to encourage formulae that differ only by commuting arguments of the addition and multiplication to be closer together in their latent representations.
Note that in their paper, for the Forward task, Lample and Charlton trained the model on 20 million examples from a dataset of 40 million samples in total. As we have carried out many experiments, and to prevent unnecessary use of resources, we have run on smaller subsets of their training data ($500k$, $1M$, $2M$ and $5M$ training examples). 

\subsection{Tree positional embeddings}

\paragraph{Related work}
Shiv and Quirk \cite{X31} proposed a one-hot positional encoding to facilitate the relationships between symbols within the tree structure in the code translation task. Later, Wang et al. \cite{X28, Wang2021} adapted this one-hot encoding approach for mathematical tasks by simplifying it to a binary encoding. However, this approach requires maintaining a stack during decoding to track tree positions and adding extra stop tokens, which increases implementation complexity. A subsequent study investigated which component, when jointly modelling natural and mathematical language, contributes most to model performance. Scarlatos and Lan \cite{Scar6Tree} merged multiple properties of a mathematical expression into an LLM. Their results suggested that tree positional embeddings account for most of the performance improvement. Unlike these previous approaches, Park et al. \cite{SymPlex} combined traversal-aware positional embedding with tree relative self-attention to process symbolic DE expressions. Since traversal-aware positional embedding and tree relative self-attention were implemented simultaneously, the individual contribution of each component remains unclear.
In our task, the encoder aims to capture the semantics of the source expressions (that is, the differential equations we wish to solve); structurally enhanced representations would facilitate this. Moreover, their data also consists of mathematical expressions.
Therefore, we adapted the method from \cite{X28, Wang2021} to our symbolic ODE tasks
by encoding the node relationships within mathematical expression trees. The tree positional embedding contains two parts: node embeddings along the tree traversal and tree positional encoding.

\paragraph{Node embeddings}
The source expressions are represented as binary trees and serialised into a sequence via prefix traversal. We obtain the node embeddings of these serialised input expressions using a learnable embedding layer. Each node in the tree is represented by a trainable embedding $x_t$ of dimension M, where $t$ is the position of the node in the serialised source expression.

\begin{figure}[htbp]
    \centering
    \includegraphics[width=0.55\textwidth]{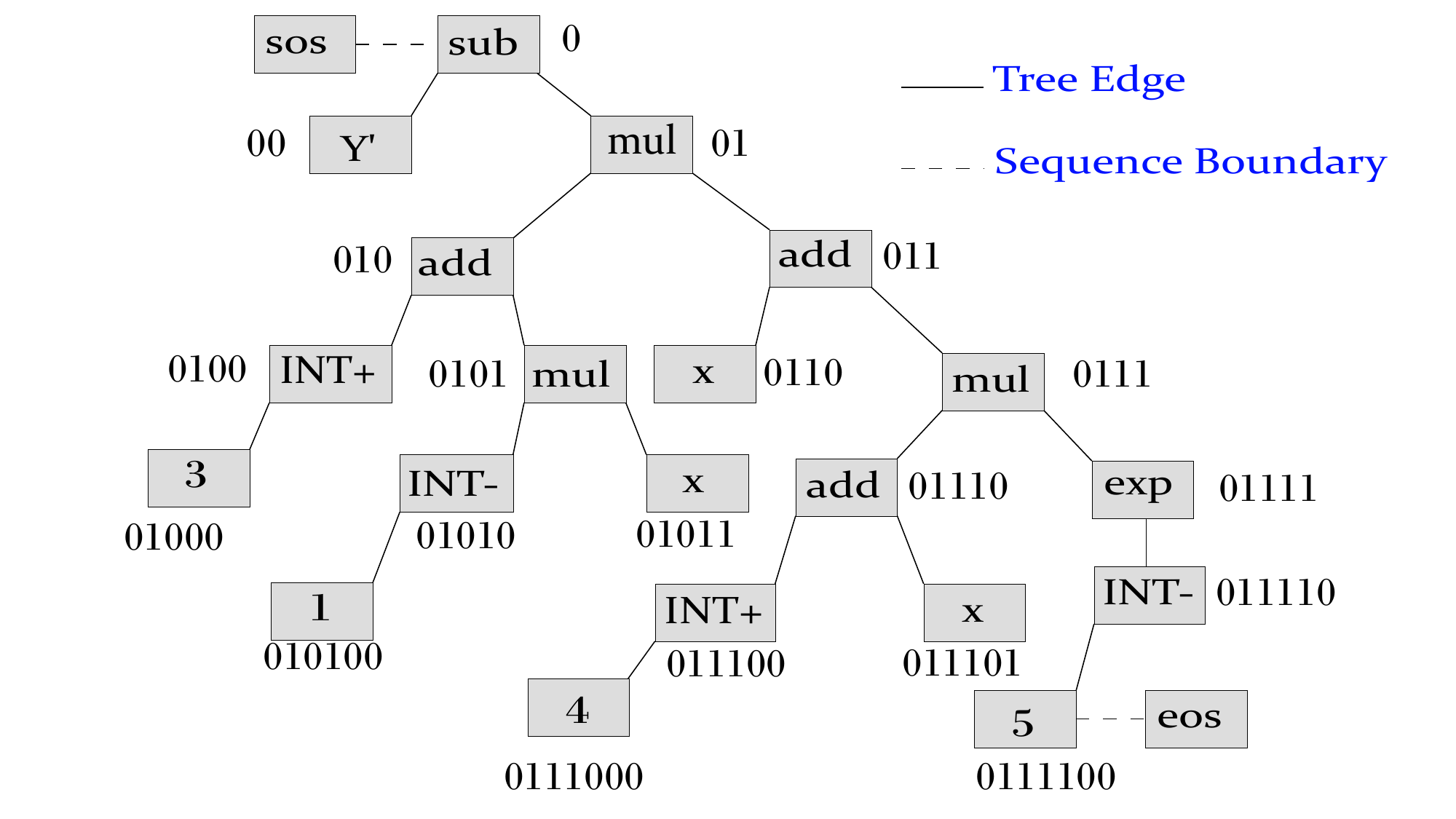}
    \caption{Example of encoding an ODE expression using tree positional encoding. Start by encoding the root node 'sub' as 0, then encode its subsequent node by appending 0 to its left node and 1 to its right node. After encoding, SOS and EOS are added at the beginning and end of the expression.}
    \label{fig:sample_tree}
\end{figure}

\paragraph{Tree positional encoding}
The encoding process starts at the top of the binary tree, following the tree prefix traversal order. We encode the root node with '0'. For each branch, the left child is encoded by appending '0' to its parent's encoding, and the right child is encoded by appending '1' after its parent's encoding. Figure \ref{fig:sample_tree} shows the mathematical expression tree's encoding process.

This encoding process yields a positional encoding list denoted by $p_d$, which cannot be used directly by the model. To convert the discrete encoding list into dense vectors, we first add 'sos' and 'eos' tokens at the beginning and end of the expression. Then we unify their lengths within a batch and leverage a learnable embedding layer to map the original positional encoding list into positional dense vectors ($p_t$). The resulting positional dense vectors and corresponding node embeddings are then fed into a bidirectional gated recurrent unit (bi-GRU), following \cite{Wang2021}. The process of fusing token embedding and positional dense vectors can be expressed as follows: 
\begin{equation}
    \operatorname{h} = \operatorname{f_c}((x_t;p_t))\\
\end{equation}
where $(x_t;p_t)$ denotes the concatenation of the $M$ dimensional node embedding, $x_t$, and the $D$ dimensional positional dense vector, $p_t$.  The bi-GRU function,
$\operatorname{f_c}: \mathbb{R}^M \times \mathbb{R}^D \rightarrow \mathbb{R}^M$, ensures that the combined embedding has the same dimensionality as the node embedding.
While in \cite{X28, Wang2021} they computed the final embedding as a weighted combination of GRU latent states, we instead add the GRU output to the corresponding node embeddings. This addition augments each node representation with its structural positional information. We only apply tree positional embedding at the encoder. This is because the encoder is responsible for capturing the semantics of the source expressions. Keeping the decoder unchanged while using absolute positional embeddings would simplify the process, eliminating the need to maintain a stack or introduce additional stop tokens.

\subsection{Contrastive Learning}

The aim of contrastive learning is to encourage the network to represent differential equations that differ only by commuting the subtrees of the addition and multiplication operators to be closer together.  Clearly, such expressions represent equivalent equations and have the same solution. To achieve this we create augmented versions of our formulae by randomly choosing to swap the children of  'mul' (the multiplication token) or 'add' (the addition token) tokens in the source expressions. We use SimCLR \cite{Chen2020SimCLR} to encourage expressions that are known to be mathematically identical to be close together in latent space. SimCLR does not require labelling the expressions or mining hard negative samples. In the SimCLR method we create a batch consisting of some source expressions $x_i$ and augmented expressions $x_j^+$. Other instances in the batch are treated as negative samples. While there is a very low likelihood that our data contains duplicate expressions, we cannot completely guarantee that there is no possibility that some mathematically equivalent expressions are treated as negatives. The loss functions can be expressed as follows
\begin{align}
    \mathrm{\ell_{i,j}} &= -\log\!\left( \frac{\exp(\mathrm{sim}(z_i, z_j^+)/\tau)}{\sum_{k=1}^{2N} \mathbf{1}_{[k \ne i]} \exp(\mathrm{sim}(z_i, z_k)/\tau)} \right)
\end{align}
\begin{align}
    \mathrm{loss\_cl} &= \sum_{i=1}^{N} \ell_{i,j}
\end{align}
\begin{align}
    \mathrm{loss} &= \mathrm{loss\_en} + \mathrm{loss\_cl}
\end{align}
where $z_i$ and $z_j^+$ are the normalised feature embedding of $x_i$ and $x_j^+$, $\tau$ is the temperature, $\mathbf{1}_{[k \ne i]} $ is an indicator function to exclude the anchor sample itself, $\ell_{i,j}$ is the loss of a pair of positive examples, $\mathrm{loss\_cl}$ is the contrastive loss within a batch and $\mathrm{loss\_en}$ is the cross-entropy loss.
When computing the contrastive loss, there are no standard pooling approaches for mathematical expressions. In the BERT architecture, the CLS token serves as the learned global representation. Due to self-attention, the first token can aggregate information from all tokens in the expression. Therefore, we follow this practice by using the first token of the source expression as its global embedding. Specifically, we extracted the first token's hidden state from the encoder's outputs. Subsequently, we applied the L2 normalisation to obtain $z_i$. The normalised results are used to compute the contrastive loss.

\section{Experimental setup}
\subsection{Data}
\begin{figure}[htbp]
  \centering   
    \includegraphics[width=0.7\textwidth]{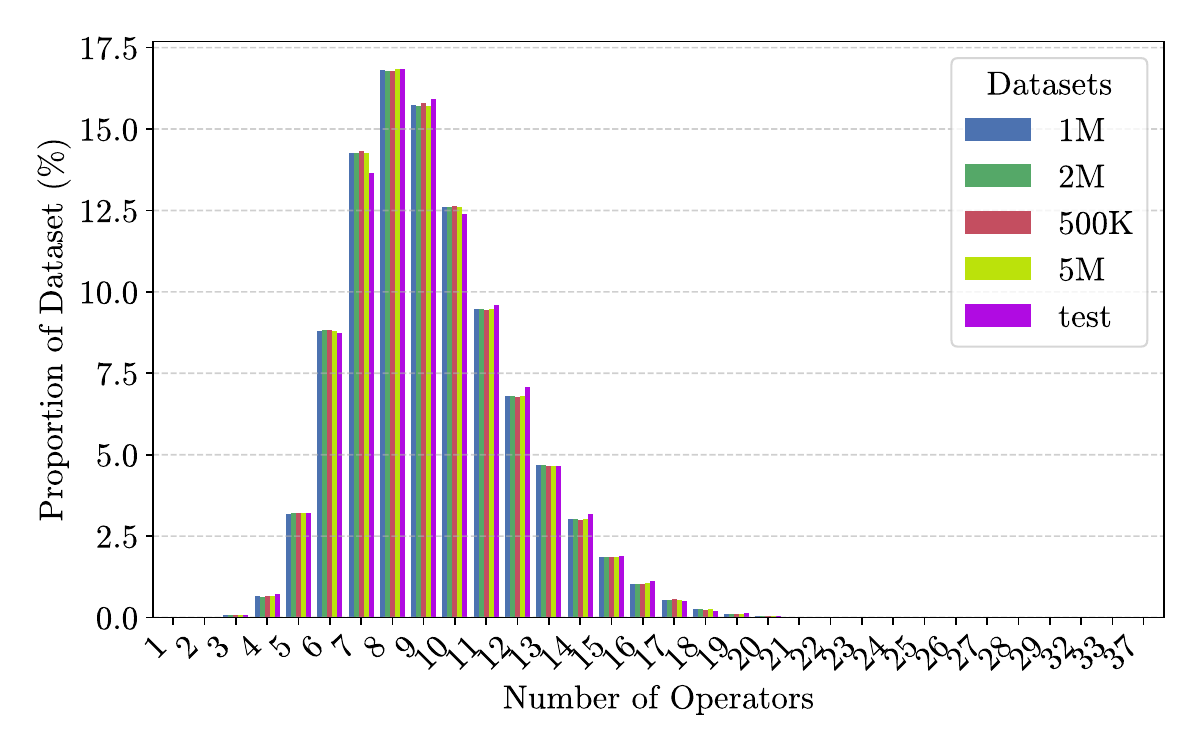}
  \caption{Training and test data distribution. The training and test data were sampled from the Forward task of training data in~\cite{ref}. The x-axis is the number of operators in the expression. The y-axis is the proportion of the corresponding category.}
  \label{fig:data_dis}
\end{figure}

We obtained the training and test data from the original training dataset of the Forward (fwd) task in \cite{ref}. This data includes approximately $40M$ samples, ensuring a diverse representation of symbolic ODE expressions. To systematically evaluate the scaling behaviour between data size and the efficacy of the proposed methods, we sampled training datasets of varying sizes: $500K$, $1M$, $2M$, and $5M$. Additionally, all models were evaluated on a fixed $10K$-sample test set with the same distribution as the training set. 
In addition to this in-distribution dataset (IDD), we also consider testing on out-of-distribution datasets. These two OOD datasets come from the Lample and Charton paper~\cite{ref}. These datasets, \textit{fwd\_test}, and \textit{ibp\_test} were generated using different scenarios from the data used for training and consequently have a different distribution (in terms of length of expressions) from the training data, as illustrated in Figure~\ref{fig:dist_ood}.  We test on these datasets to assess generalisation performance under domain shift.
Using IDD data ensures that the model may generalise within the target functional domain and provides an unbiased estimate of performance on the test data. Thus, we can easily isolate the specific advantages provided by the techniques used rather than the noise from data variance. Figure~\ref{fig:data_dis} shows the data distributions for the training and the test datasets. 

For contrastive learning, we generated positive data by swapping the operands of the commutable operators (i.e., 'add' or 'mul' token) in the expression tree. There can be multiple commutable operators in one expression. To reduce learning complexity, we execute the swapping mechanism for each expression only once, when commutable operators are encountered in order. If an expression contains no commutable operators, the original expression is used as the positive sample. Such instances account for $0.0036\%$ of the total dataset ($500K$), thus their impact on model training is insignificant. Consistent with standard practice, we treat other data in the same batch as negative samples.

\subsection{Model}
We use the encoder-decoder transformer. Model parameter choices follow \cite{ref}. The encoder and decoder have 6 layers, respectively. Each layer has 8 attention heads. The embedding dimension is 512. We train a transformer model with absolute position embedding (APE) and tree positional embedding (TPE) separately,
Optimisation was performed using an Adam optimiser with an initial learning rate of 0.0001 and weight decay of 0.0001 in a batch size of 256. The learning rate was modulated by a cosine annealing learning rate scheduler at a frequency of every 10 epochs. Then, we implemented instance-level contrastive learning after an initial warm-up phase. For all contrastive learning-related experiments, we first train the model with cross-entropy for the first 10 epochs. This allows the model to learn task-relevant representations before applying the contrastive objective. Then we train the model using cross-entropy and contrastive losses from the 11th epoch through the end of training. The temperature coefficient is set to 1 when computing contrastive loss. To maintain a controlled experimental environment and facilitate a clear comparison of scaling behaviours, all experiments were trained for a fixed 60 epochs. The choice of 60 is that training had converged by this stage in preliminary experiments. This avoids other factors, such as the variations during training caused by early-stopping criteria. We use exact match as a metric, comparing the generated solution against its ground truth using Sympy. 

\section{Results}
\subsection{Tree positional embedding}
\begin{figure}[htbp]
  \centering  
 
    \includegraphics[width=0.7\textwidth]{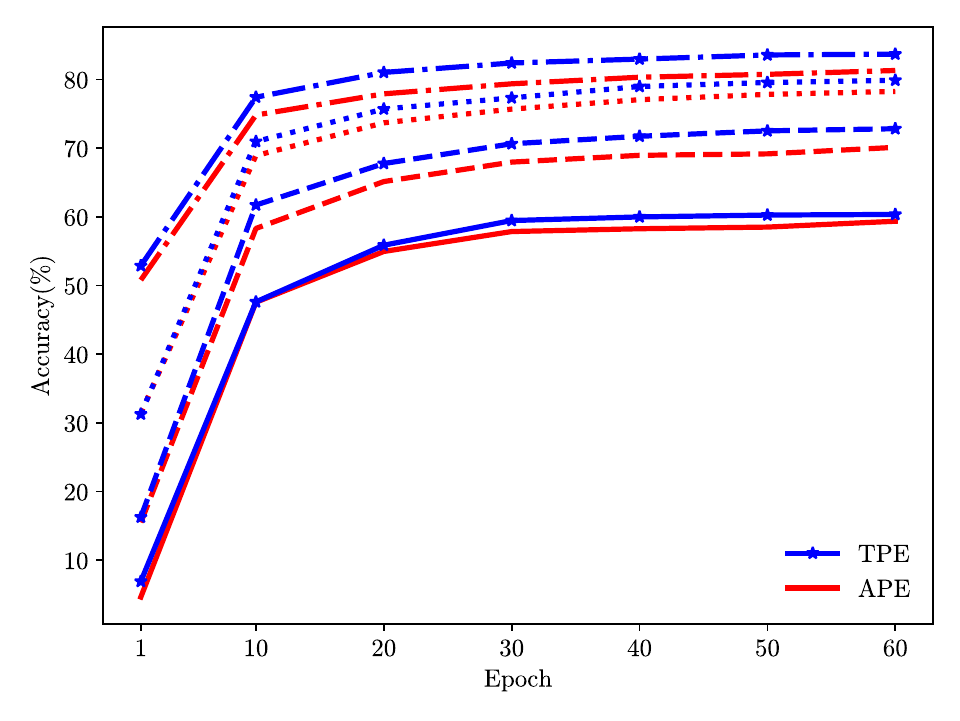}

  \caption{Performance of the model using absolute positional embedding (APE) and the model using tree position embedding (TPE) trained on $500K$ (-), $1M$ (- -), $2M$(..) and $5M$ (-.) data. }
  \label{fig:tree_basic_all}
\end{figure}
Figure \ref{fig:tree_basic_all} shows the empirical results for models with identical configurations that are trained with absolute position embeddings (APE) and tree positional embeddings (TPE). These results suggest that encoding structural information is beneficial for a model to solve symbolic ODE problems. It is evident that the performance gap between these two models does not scale consistently with the size of the training data. The performance gap first increased from $1\%$ at $500K$ samples to $2.7\%$ at $1M$ samples, then decreased to $1.6\%$ at $2M$ samples, before rising again to $2.4\%$ at $5M$ samples.  Some of this variation is clearly due to random fluctuations (we would expect the size of fluctuations on our test set of 10K samples to be around 0.5\%), but there may also be a systematic pattern.
One possible explanation is that, as the training set grows and reaches a certain amount, i.e., $2M$. \textit{Model\_APE} becomes increasingly capable of fitting patterns in the training data, thereby reducing the relative advantage provided by TPE. But when the data size increases to $5M$, the performance gap rebounds slightly. It suggests that the structural information introduced by TPE does not vanish entirely at larger data scales. This can be supported by plotting their performance gap against the number of operators per expression, as shown in Figures \ref{fig:diff1}-\ref{fig:diff4}.
Another observation is that at the first training epoch, \textit{Model\_TPE} achieved slightly higher performance than \textit{Model\_APE}. This performance gap continues to increase as the training epoch grows. This shows that explicitly imposing structural information improves model learning efficiency.
Overall, the parent-child relationship encoded via TPE enables the model to perform better and more consistently across different data sizes.

\subsection{Generalisation to Out of Distribution Data and Ablation Study}
We train a transformer model with identical configurations using $500K$ data points across different experimental settings, as shown in Table \ref{tab:500K_all_zoomin}. We then evaluate it on IDD and OOD datasets, respectively.  We performed an ablation study where we considered models using absolute positional embeddings (APE), tree positional embeddings (TPE), with or without contrastive learning (CL) and where we also considered combining the original dataset with the augmented dataset (Merge).

The results in Table \ref{tab:500K_all_zoomin} show that contrastive learning slightly increased the model's performance by $0.3\%$ when comparing the results of \textit{Model\_TPE} and \textit{Model\_TPE\_CL}. However, the results from \textit{Model\_APE} and \textit{Model\_APE\_CL} suggest that contrastive learning did not help the model to capture meaningful invariance representations of these expressions with APE. Instead, it made \textit{Model\_APE\_CL} underperform \textit{Model\_APE}, as the performance dropped by $1\%$. It might be that \textit{Model\_APE\_CL}, trained with APE, does not explicitly encode the structure of symbolic expressions. Contrastive pairs generated through commutative transformations may be more difficult for \textit{Model\_APE\_CL} to align in the latent space. Even though the marginal performance gain of \textit{Model\_TPE\_CL} is against \textit{Model\_TPE}. This result shows that TPE better preserves expression equivalence for structural invariance during contrastive learning, which can be supported by the T-SNE visualisation of their latent embeddings in Figure \ref{fig:t-sne}. The comparison of the latent embeddings of the same pair of positive expressions between \textit{Model\_TPE\_CL} and \textit{Model\_APE\_CL} suggests that TPE enables contrastive learning. 
For models that apply TPE, directly conducting supersized training on augmented data (\textit{Model\_TPE\_merge}) outperforms training with contrastive loss (\textit{Model\_TPE\_CL}). This suggests that the learning signal from contrastive learning might not be as strong as that from direct supervised learning in the IDD settings. 
Additionally, we observed that the model with APE trained on merged data (\textit{Model\_APE\_merge}) performs slightly worse than \textit{Model\_TPE}. That is to say, \textit{Model\_TPE} trained only on raw data ($500K$) outperforms \textit{Model\_APE\_merge} trained on merged data ($1M$). This indicates an additional advantage: TPE can yield larger gains than naive data scaling of APE, which can be interpreted as TPE being more data-efficient than APE.
\begin{table}
    \centering
    \begin{tabular}{|l|c|}
    \hline
        Model & Accuracy($\%$)\\\hline        
        \textit{Model\_APE} & 59.4\\ \hline
        \textit{Model\_TPE} & 60.4\\ \hline
        \textit{Model\_TPE\_CL} & 60.7\\ \hline
        \textit{Model\_APE\_merge} & 60.3\\ \hline       
        \textit{Model\_APE\_CL} & 58.4\\ \hline
        \textit{Model\_TPE\_merge} & \textbf{62.4}\\ \hline
    \end{tabular}
    \caption{The evaluation results of models with different configurations trained with 500K data. APE: Absolute positional embedding; TPE: Tree positional embedding; CL: Contrastive learning; Merge: The model was trained with raw and augmented training data. The best result is shown in bold. }
    \label{tab:500K_all_zoomin}
\end{table}

\begin{table}
    \centering
    \begin{tabular}{|l|c|c|}
    \hline
        Model & \textit{fwd\_test}($\%$)& \textit{ibp\_test}($\%$)\\\hline        
        \textit{Model\_APE} & 64.5&  62.6\\ \hline
        \textit{Model\_TPE} & 66.3&  63.8\\ \hline
        \textit{Model\_TPE\_CL} & 66.6&  64.3\\ \hline
        \textit{Model\_APE\_merge} & 65.2& 63.7\\ \hline       
        \textit{Model\_APE\_CL} & 63.9& 62.4\\ \hline
        \textit{Model\_TPE\_merge}&\textbf{67.0} &\textbf{64.6}\\ \hline
    \end{tabular}
    \caption{The accuracy of \textit{Model\_APE}, \textit{Model\_TPE}, \textit{Model\_TPE\_CL}, \textit{Model\_APE\_merge}, \textit{Model\_APE\_CL} and \textit{Model\_TPE\_merge} trained with 500K data evaluated on fwd and ibp tasks. The best results are shown in bold.}
    \label{tab:500K_ood_task}
\end{table}

The performance improvements observed in the IDD setting are preserved under distribution shift, as shown in Table \ref{tab:500K_ood_task}. Based on these results, models using TPE outperform those using APE. \textit{Model\_TPE\_merge}, trained with TPE on both raw and augmented data, achieved the best performance among these models and also demonstrated that explicit exposure to augmented data may be more beneficial than leveraging a contrastive loss to learn representation invariance in OOD. In the IDD setting, the performance gap between \textit{Model\_TPE\_merge} and \textit{Model\_TPE\_CL} is 1.8 $\%$, but decreases to 0.4 $\%$ in both the \textit{fwd\_test} and \textit{ibp\_test} tasks.
This suggests that the benefit of direct supervision on augmented data is more pronounced under in-distribution conditions, where structural information might be fully exploited to learn the pattern during training. In contrast, under distribution shift, generalisation limits might primarily constrain performance, and the advantage introduced by direct supervision on augmented data becomes less significant. The reduced performance gap suggests that contrastive learning might help stabilise performance under OOD conditions. In general,
\textit{Model\_TPE\_merge} and \textit{Model\_TPE\_CL} achieved the best and second-best performance, respectively, in both IDD and OOD settings. This consistent ranking suggests that while distribution shift reduces the magnitude of improvements, it does not alter TPE's comparative advantage. In other words, TPE provides a relatively stable and non-degrading performance gain that remains beneficial across different evaluation regimes.

\section{Conclusions}
We investigated the utility of tree positional embedding and contrastive learning for solving symbolic ODEs. The results suggest that tree positional embedding improves model performance and maintains a relatively strong advantage across different data sizes and tasks. Additionally, we found that the structural information provided by tree positional embedding is the foundation to enable contrastive learning in this task. The effectiveness of contrastive learning in learning invariant representations of a pair of symbolic expressions is limited compared to explicitly exposing the model to these augmentations. Several factors may contribute to this observation. 
The relatively small batch sizes imposed by computational constraints may reduce the effectiveness of contrastive learning, while applying a single transformation per expression may be insufficient to capture the rich semantic invariances inherent in complex symbolic structures. More broadly, these findings suggest that supervised training with
augmented data may be more effective for symbolic ODE tasks than contrastive learning for invariant representations.

However, our work has limitations. We conducted experiments on a single transformer model type and trained on one task. This is because we specifically selected this model to provide a direct comparison with the original baseline.
Further work may explore diverse datasets and transformer models. When computing contrastive loss, we used nominalised embeddings to compute contrastive loss at a relative stable training stage. Therefore, we did not tune temperature's value. We did not balance the cross entropy and contrastive losses either due to the empirically stable training, instead we adopted equal weighting to avoid additional parameter tuning. A comprehensive study of the influence of alpha and temperature is left for future work.

This work was supported by the UK Research and Innovation Centre for Doctoral Training in Machine Intelligence for Nano-electronic Devices and Systems [EP/S024298/1].
\bibliographystyle{ieeetr}
\bibliography{references}

\clearpage
\appendix
\section{Supplementary Figures}

\label{appendix:figures}

\setcounter{figure}{0}
\renewcommand{\thefigure}{A.\arabic{figure}}

\begin{figure}[htbp]
    \centering
    \includegraphics[width=0.7\linewidth]{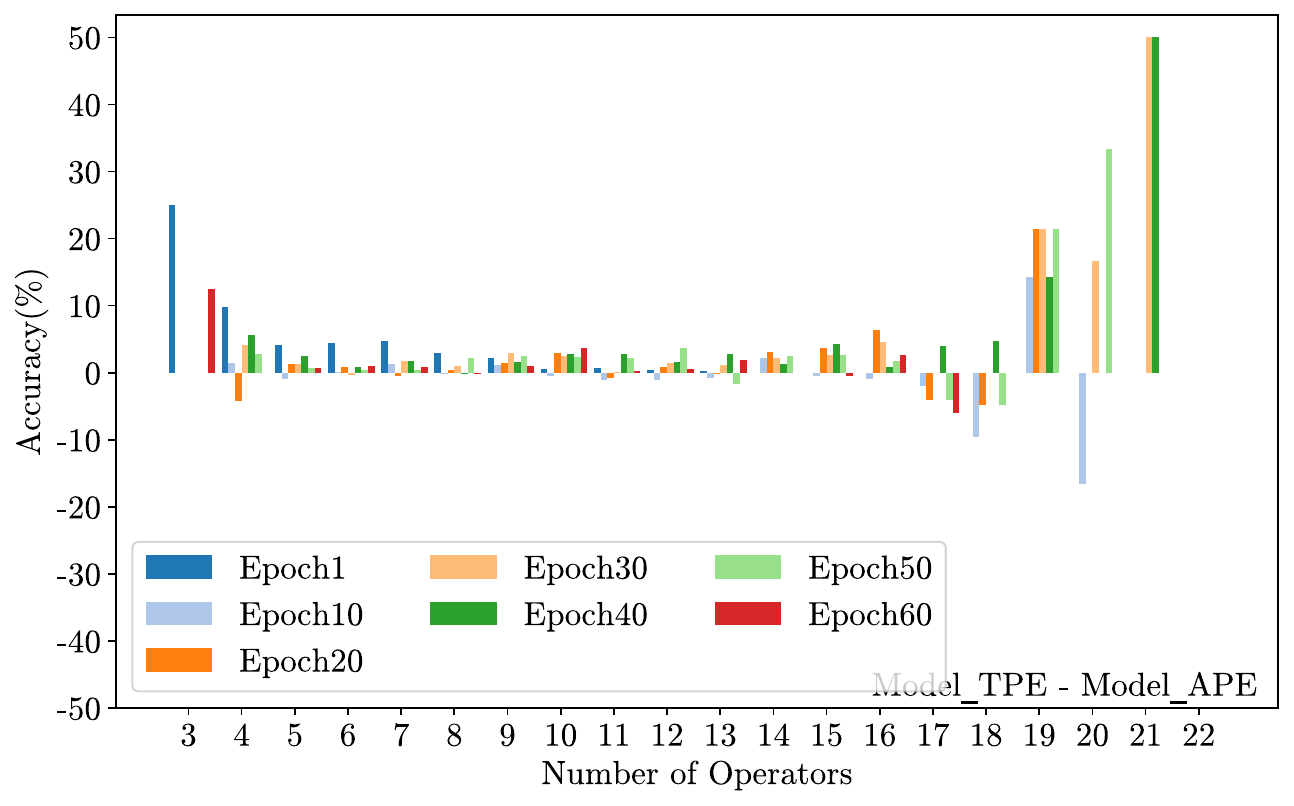}
    \caption{Performance difference between \textit{Model\_TPE} and \textit{Model\_APE} trained with $500K$ data.}
    \label{fig:diff1}
\end{figure}

\begin{figure}[htbp]
    \centering
    \includegraphics[width=0.7\linewidth]{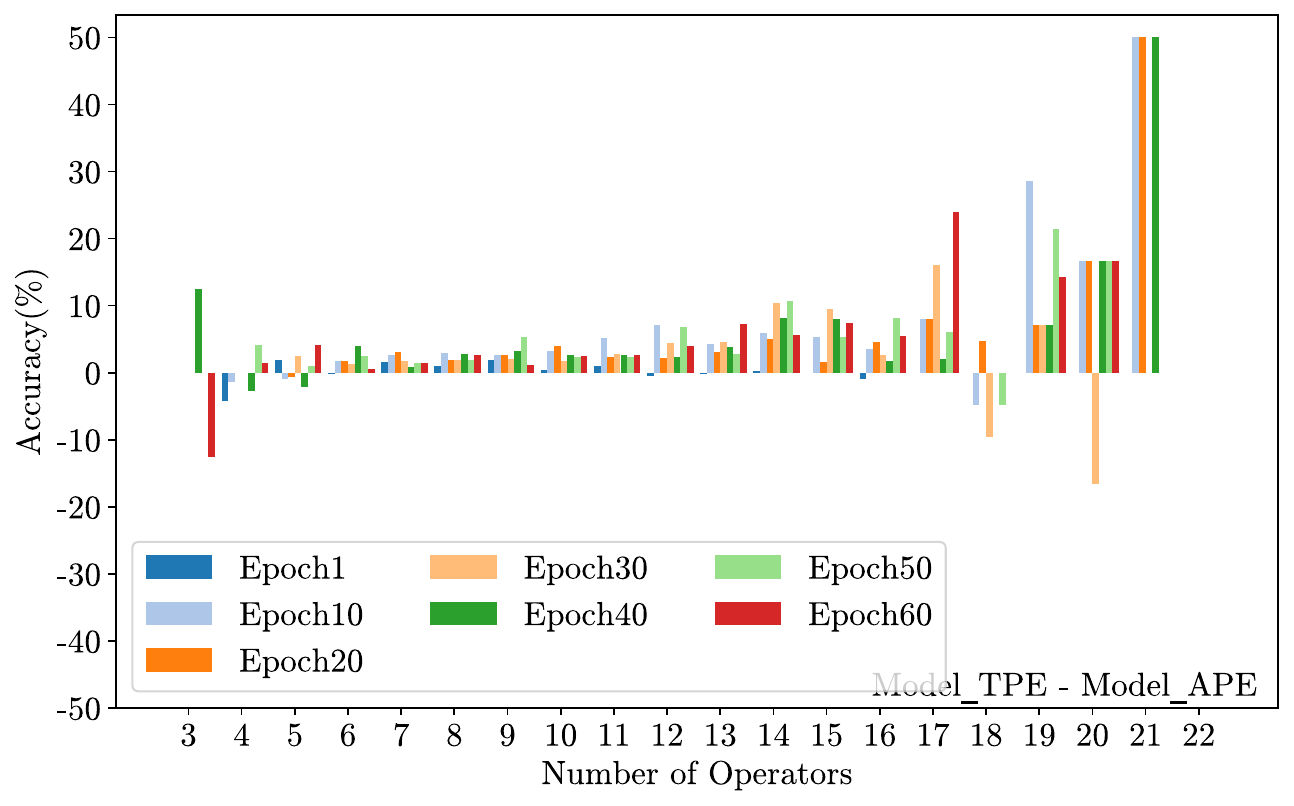}
    \caption{Performance difference between \textit{Model\_TPE} and \textit{Model\_APE} trained with $1M$ data.}
    \label{fig:diff2}
\end{figure}

\begin{figure}[htbp]
    \centering
    \includegraphics[width=0.7\linewidth]{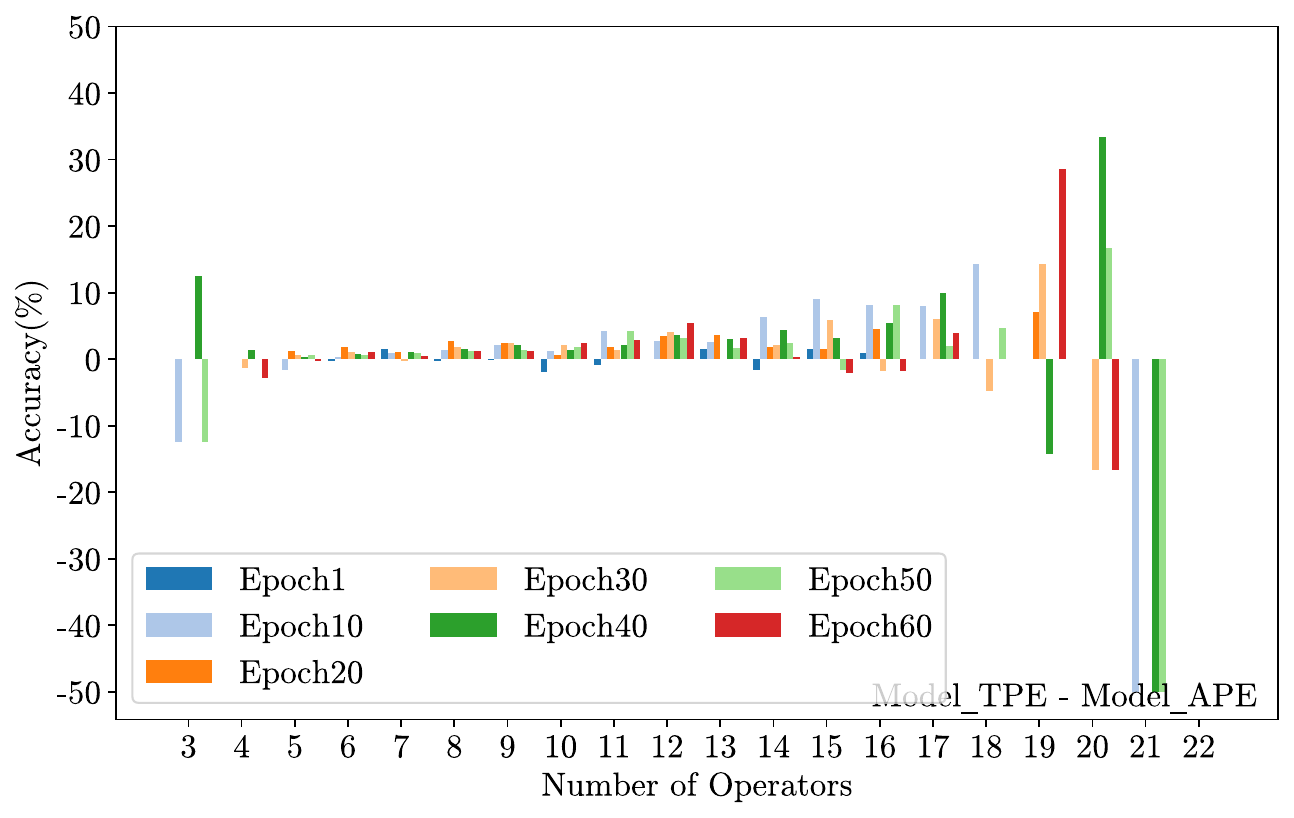}
    \caption{Performance difference between \textit{Model\_TPE} and \textit{Model\_APE} trained with $2M$ data.}
    \label{fig:diff3}
\end{figure}

\begin{figure}[htbp]
    \centering
    \includegraphics[width=0.7\linewidth]{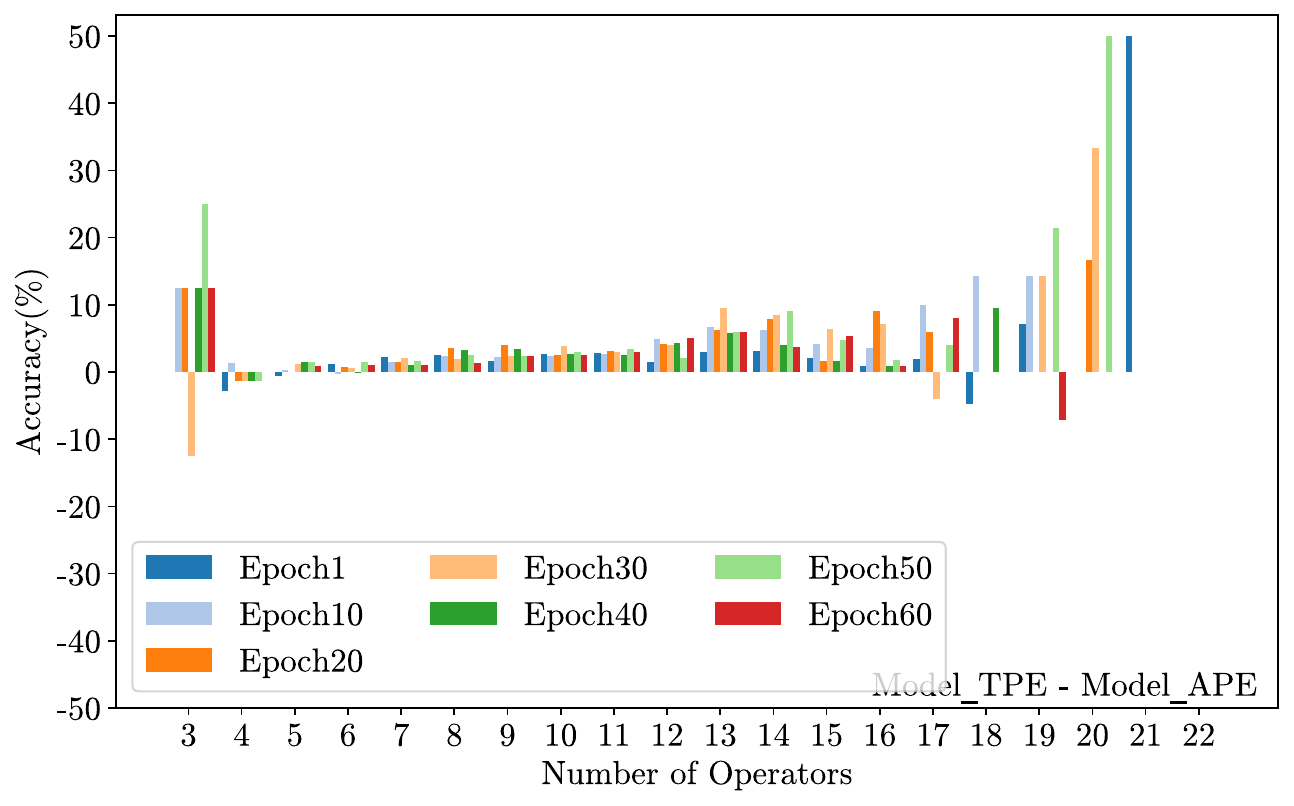}
    \caption{Performance difference between \textit{Model\_TPE} and \textit{Model\_APE} trained with $5M$ data.}
    \label{fig:diff4}
\end{figure}

\begin{figure}[htbp]
  \centering  
 
    \includegraphics[width=\textwidth]{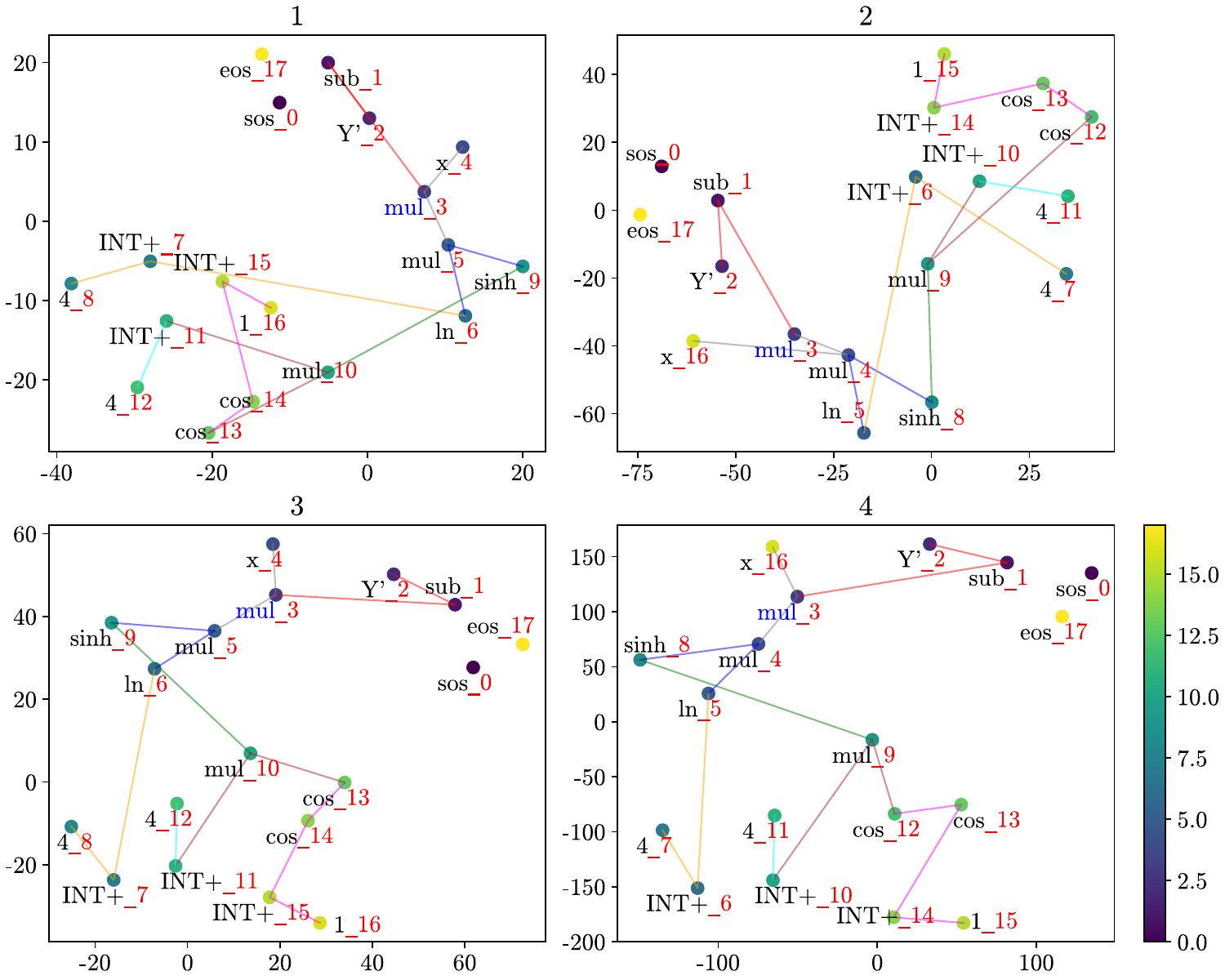}

  \caption{T-SNE visualisation of latent embeddings for the original expression (the left side) and its augmented expression (the right side). Nodes are denoted by their name, an underline, and their position (marked as red) in the serialised expression. The node marked as blue is the commutable operator to generate the augmented data. Edges of the same colour connect nodes within the same subtree. SOS and EOS do not belong to any subtree. The top row shows the results of \textit{Model\_APE\_CL}, and the bottom row shows the results of \textit{Model\_TPE\_CL}.}
  \label{fig:t-sne}
\end{figure}

\begin{figure}[htbp]
    \centering
    \includegraphics[width=0.7\linewidth]{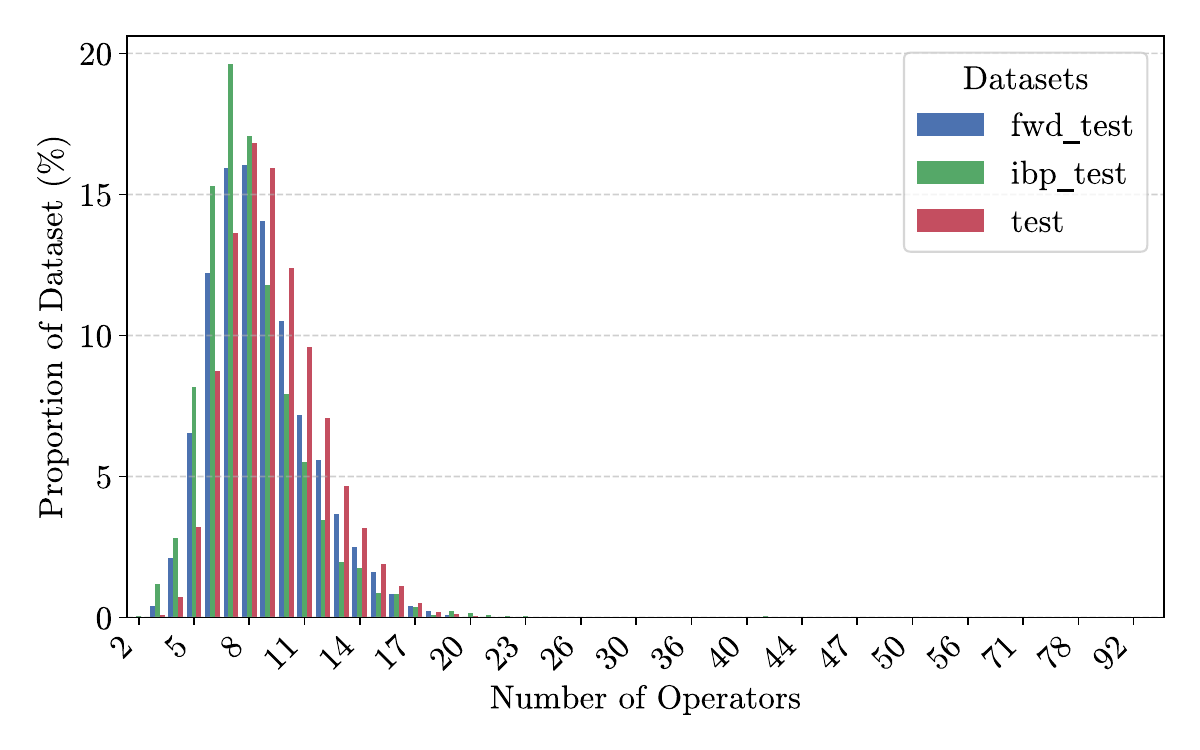}
    \caption{Distribution of test, \textit{fwd\_test} and \textit{ibp\_test} datasets. The x-axis is the number of operators in the expression. The y-axis is the proportion of the corresponding category.}
    \label{fig:dist_ood}
\end{figure}
\end{document}